\documentclass[11pt,a4paper]{article}
\usepackage{times,latexsym}
\usepackage{url}
\usepackage[T1]{fontenc}
\pdfoutput=1 

\usepackage{adjustbox}

\usepackage[acceptedWithA]{tacl2021v1}

\usepackage{xspace,mfirstuc,tabulary}

\newif\iftaclinstructions
\taclinstructionsfalse 
\iftaclinstructions
\renewcommand{\confidential}{}
\renewcommand{\anonsubtext}{(No author info supplied here, for consistency with
TACL-submission anonymization requirements)}
\newcommand{\instr}
\fi

\iftaclpubformat 

\else

\fi

\pdfoutput=1

\usepackage{tikzit}

\usepackage{times}
\usepackage{caption}
\usepackage{latexsym}
\usepackage{subfigure}

\usepackage{longtable}
\usepackage[T1]{fontenc}

\usepackage[utf8]{inputenc}

\usepackage{microtype}

\usepackage{paralist}

\usepackage{enumitem}
\usepackage{xspace}

\usepackage{booktabs} 

\usepackage{wasysym} 
\usepackage{array}
\usepackage{multirow}

\usepackage{tikz}
\usepackage{pgfplots}
\usepackage{float}
\usepackage{arydshln}
\usepackage{centernot}
\usepackage{bbm}
\usepackage[normalem]{ulem}

\tikzstyle{new style 0}=[fill=white, draw={rgb,255: red,173; green,175; blue,191}, shape=circle, minimum width=.5cm, ultra thick]
\tikzstyle{new style 1}=[fill=white, draw=black, shape=rectangle]
\tikzstyle{new style 2}=[fill=white, draw=red, shape=rectangle]

\tikzstyle{new edge style 0}=[->, fill=none, draw=red, ultra thick]
\tikzstyle{new edge style 1}=[<-, fill=none, ultra thick]
\tikzstyle{new edge style 2}=[draw=black, fill=none, dashed, <-, ultra thick]
\tikzstyle{new edge style 3}=[<-, ultra thick]
\tikzstyle{new edge style 4}=[->, ultra thick]
\tikzstyle{new edge style 5}=[draw=black, fill=none, dashed, ->, ultra thick]

\usepackage{tabulary}

\newcommand{\skippedDetails}[1]{}

\usepackage{xcolor}

\definecolor{orange}{rgb}{1,0.5,0}
\definecolor{mdgreen}{rgb}{0.05,0.6,0.05}
\definecolor{mdblue}{rgb}{0,0,0.7}
\definecolor{dkblue}{rgb}{0,0,0.5}
\definecolor{dkgray}{rgb}{0.3,0.3,0.3}
\definecolor{slate}{rgb}{0.25,0.25,0.4}
\definecolor{gray}{rgb}{0.5,0.5,0.5}
\definecolor{ltgray}{rgb}{0.7,0.7,0.7}
\definecolor{purple}{rgb}{0.7,0,1.0}
\definecolor{lavender}{rgb}{0.65,0.55,1.0}

\newcommand{\alphaOne}{$\alpha_1$\xspace}
\newcommand{\alphaTwo}{$\alpha_2$\xspace}
\newcommand{\alphaThree}{$\alpha_3$\xspace}
\newcommand{\trainSet}{$Train$\xspace}

\newcommand{\rowKey}[1]{\textit{#1}\xspace}
\newcommand{\entaillabel}[1]{\textcolor{mdgreen}{\textsc{#1}}\xspace}
\newcommand{\entail}{\entaillabel{Entail}}
\newcommand{\contradict}{\entaillabel{Contradict}}
\newcommand{\neutral}{\entaillabel{Neutral}}

\def\boxit#1{%
  \smash{\color{red}\fboxrule=1pt\relax\fboxsep=2pt\relax%
  \llap{\rlap{\fbox{\vphantom{0}\makebox[#1]{}}}~}}\ignorespaces
}

\newcommand{\datasetName}{{\sc InfoTabS}\xspace}

\usepackage{amssymb}
\usepackage{pifont}
\newcommand{\cmark}{\ding{51}}%
\newcommand{\xmark}{\ding{55}}%

\newcommand{\quash}[1]{}

\newcommand{\theTitle}{Is My Model Using the Right Evidence? Systematic Probes for Examining Evidence-Based Tabular Reasoning}

\title{\theTitle}

\author{
Vivek Gupta\textsuperscript{\rm 1},
Riyaz A. Bhat\textsuperscript{\rm 2},
Atreya Ghosal\textsuperscript{\rm 3},\\
\bf 
Manish Shrivastava\textsuperscript{\rm 3},
Maneesh Singh\textsuperscript{\rm 2},
Vivek Srikumar\textsuperscript{\rm 1}\\
\textsuperscript{\rm 1}University of Utah,
\textsuperscript{\rm 2}Verisk Inc.,
\textsuperscript{\rm 3}IIIT-Hyderabad,\\
\{vgupta, svivek\}@cs.utah.edu, \{riyaz.bhat, maneesh.singh\}@verisk.com,\\ \{atreyee.ghosal@research, m.shrivastava@\}iiit.ac.in \\
}

\begin{document}

\maketitle

\begin{abstract}
Neural models command state-of-the-art performance across NLP tasks, including ones involving ``reasoning''. Models claiming to reason about the evidence presented to them should attend to the correct parts of the input avoiding spurious patterns therein, be self-consistent in their predictions across inputs, and be immune to biases derived from their pre-training in a nuanced, context-sensitive fashion. {\em Do the prevalent *BERT-family of models do so?} In this paper, we study this question using the problem of reasoning on tabular data. Tabular inputs are especially well-suited for the study---they admit systematic probes targeting the properties listed above. Our experiments demonstrate that a RoBERTa-based model, representative of the current state-of-the-art, fails at reasoning on the following counts: it  (a) ignores relevant parts of the evidence, (b) is over-sensitive to annotation artifacts, and (c) relies on the knowledge encoded in the pre-trained language model rather than the evidence presented in its tabular inputs. Finally, through inoculation experiments, we show that fine-tuning the model on perturbed data does not help it overcome the above challenges.
\end{abstract}

\section{Introduction}
\label{sec:introduction}
The problem of understanding tabular or semi-structured data is a challenge for modern NLP. Recently, \citet{chen2019tabfact} and \citet{gupta-etal-2020-infotabs} have framed this problem as a natural language inference question~\cite[NLI,][\emph{inter alia}]{dagan2013recognizing,snli:emnlp2015} via the TabFact and the \datasetName~datasets respectively. The tabular version of the NLI task seeks to determine whether a \emph{tabular} premise entails or contradicts a textual hypothesis, or is unrelated to it.

One strategy for such tabular reasoning tasks relies on the successes of contextualized representations~\cite[e.g.,][]{devlin2019bert,liu2019roberta} for the sentential version of the problem. Tables are flattened into artificial sentences using heuristics to be processed by these models. Surprisingly, even this na\"ive strategy leads to high predictive accuracy, as shown not only by the introductory papers but also by related lines of recent work~\cite[e.g.,][]{eisenschlos2020understanding,yin-etal-2020-tabert}.

In this paper, we ask: \emph{Do these seemingly accurate models for tabular inference effectively use and reason about their semi-structured inputs?} While ``reasoning'' can take varied forms, a model that claims to do so should at least ground its outputs on the evidence provided in its inputs. Concretely, we argue that such a model should 
\begin{inparaenum}[(a)]
\item be self-consistent in its predictions across controlled variants of the input, 
\item use the evidence presented to it, and the right parts thereof, and,
\item avoid being biased \emph{against} the given evidence by knowledge encoded in the pre-trained embeddings.
\end{inparaenum}

Corresponding to these three properties, we identify three dimensions to evaluate a tabular NLI system: robustness to annotation artifacts, relevant evidence selection, and robustness to counterfactual changes. We design systematic probes that exploit the semi-structured nature of the premises. This allows us to semi-automatically construct the probes and to unambiguously define the corresponding expected model response. These probes either introduce controlled edits to the premise or the hypothesis, or to both, thereby also creating counterfactual examples. Experiments reveal that despite seemingly high test set accuracy, a model based on RoBERTa~\cite{liu2019roberta}, a good representative of BERT derivative models, is far from being reliable. Not only does it ignore relevant evidence from its inputs, it also relies excessively on annotation artifacts, in particular the sentence structure of the hypothesis, and pre-trained knowledge in the embeddings. Finally, we found that attempts to inoculate the model~\cite{liu2019inoculation} along these dimensions degrades its overall performance.

The rest of the paper is structured as follows. \S\ref{sec:tnli} introduces the Tabular NLI task, while \S\ref{sec:reasoning} articulates the need for probing evidence-based tabular reasoning in extant high-performing models. \S\ref{sec:probe2}-\S\ref{sec:probe3} detail the probes designed for such an examination and the results thereof while \S\ref{sec:inoculation} analyzes the impact of inoculation to aforementioned challenges through model fine-tuning. \S\ref{sec:discussion} presents the  main takeaways and contextualization in the related art. \S\ref{sec:conclusion} provides concluding remarks and indicates future directions of work.\footnote{The dataset and the scripts used for our analysis are available at \url{https://tabprobe.github.io}.}
\section{Preliminaries: Tabular NLI}
\label{sec:tnli}

Tabular natural language inference is a task similar to standard NLI in that it examines if a natural language hypothesis can be derived from the given premise. Unlike standard NLI, where the evidence is presented in the form of sentences, the premises in tabular NLI are semi-structured tables that may contain both text and data. 

\paragraph{Dataset} Recently, datasets such as TabFact~\cite{chen2019tabfact} and \datasetName~\cite{gupta-etal-2020-infotabs}, and also shared tasks such as SemEval 2021 Task 9~\cite{semeval_2021:task9} and FEVEROUS~\cite{aly2021feverous}, have sparked interest in tabular NLI research. In this study, we use the \datasetName dataset for our investigations.

\datasetName consists of $23,738$ premise-hypothesis pairs, whose premises are based on Wikipedia infoboxes. Unlike TabFact, which only contains \entail and \contradict hypotheses, \datasetName also includes \neutral ones. Figure~\ref{fig:example} shows an example table from the dataset with four hypotheses, which will be our running example. 

The dataset contains $2,540$ distinct infoboxes representing a variety of domains. All hypotheses were written and labeled by MTurk workers. The tables contain a \emph{title} and two columns, as shown in the example. Since each row takes the form of a key-value pair, we will refer to the elements in the left column as the \emph{keys}, and the right column provides the corresponding \emph{values}. 

In addition to the usual train and development sets, \datasetName includes three test sets, \alphaOne, \alphaTwo and \alphaThree. The \alphaOne set represents a standard test set that is both topically and lexically similar to the training data. In the \alphaTwo set, hypotheses are designed to be lexically adversarial, and the \alphaThree tables are drawn from topics not present in the training set. We will use all three test sets for our analysis.

\noindent\begin{minipage}{0.96\columnwidth}
\bigskip
  \centering
  {
  \footnotesize
  \begin{center}
    \begin{tabular}{>{\raggedright}p{0.25\linewidth}p{0.6\linewidth}}
      \toprule
      \multicolumn{2}{c}{\bf Breakfast in America}                                                    \\
      \midrule
      {\bf Released	} & 	29 March 1979 \\
                                              
      {\bf Recorded} & May–December 1978                                                  \\ 
      {\bf Studio}           &  The Village Recorder (Studio B) in Los Angeles          \\  
      {\bf Genre}                  & pop ; art rock ; soft rock                             \\ 
      {\bf Length}      & 46:06                                         \\  
     {\bf Label}             & A$\&$M                                                 \\ 
     {\bf Producer} & 	Peter Henderson, Supertramp\\
      \bottomrule
    \end{tabular}
  \end{center}
}
  {\footnotesize
    \begin{enumerate}[nosep]
    \item[H1:] Breakfast in America is a pop album with a length of 46 minutes.
    \item[H2:] Breakfast in America was released at the end of 1979.
    \item[H3:] Most of Breakfast in America was recorded in the last month of 1978.
    \item[H4:] Breakfast in America has 6 tracks.
    \end{enumerate}
  }
    \captionof{figure}{\small An example of a tabular premise from \datasetName. The hypotheses H1 is entailed by it, H2 contradicts it, and H3, H4 are neutral, i.e., neither entailed nor contradictory.}
  \label{fig:example}  
\end{minipage}

\paragraph{Models over Tabular Premises}
Unlike standard NLI, which can use off-the-shelf pre-trained contextualized embeddings, the semi-structured nature of premises in tabular NLI necessitates a different modeling approach.

Following~\citet{chen2019tabfact}, tabular premises are flattened into token sequences that fit the input interface of such models. While different flattening strategies exist in the literature, we adopt the \emph{Table as a Paragraph} strategy of~\citet{gupta-etal-2020-infotabs}, where each row is converted to a sentence of the form ``The \texttt{key} of \texttt{title} is \texttt{value}''. This seemingly na\"ive strategy, with RoBERTa-large embeddings (RoBERTa$_L$ henceforth), achieved the highest accuracy in the original work, shown in Table~\ref{tab:pararesults}.\footnote{Other flattening strategies have similar performances \citep{gupta-etal-2020-infotabs}.} The table also shows the hypothesis-only baseline \cite{poliak2018hypothesis, gururangan2018annotation} and human agreement on the labels.\footnote{Preliminary experiments on the development set showed  that RoBERTa$_L$ outperformed other pre-trained embeddings. We found that BERT$_{B}$, RoBERTa$_{B}$, BERT$_{L}$, ALBERT$_{B}$ and ALBERT$_{L}$ reached development set accuracies of 63.0$\%$, 67.23$\%$, 69.34$\%$, 70.44$\%$ and 70.88$\%$, respectively. While we have not replicated our experiments on these other models due to a prohibitively high computational cost, we expect the conclusions to carry over to these other models as well.}

To study the stability of the models to variations in the training data, we performed  5-fold cross validation (5xCV). An average cross validation accuracy of 73.53$\%$ with a standard deviation of 2.73$\%$ was observed on the training set which is close to the performance on the \alphaOne test set (74.88\%). In addition, we also evaluated performance on the development and test sets. The penultimate row of Table~\ref{tab:pararesults} presents the performance for the model trained on the entire training data, while the last row presents the performance of the 5xCV models. The results demonstrate that model performance is reasonably stable to variations in the training set.

\noindent\begin{minipage}{\columnwidth}
\bigskip
\small
\begin{tabular}{p{1.3cm}p{1cm}p{1cm}p{1cm}p{1cm}}
\toprule
Model & dev & \alphaOne & \alphaTwo & \alphaThree \\
\midrule
Human  &\bf 79.78 & \bf 84.04 &\bf 83.88 &\bf 79.33 \\
Hypothesis Only & 60.51 & 60.48 & 48.26 & 48.89 \\
RoBERTa$_{L}$ & 75.55 & 74.88 & 65.55 & 64.94 \\ 
5xCV & 73.59$_{(2.3)}$ & 72.41$_{(1.4)}$ & 63.02$_{(1.9)}$ & 61.82$_{(1.4)}$ \\
\bottomrule
\end{tabular}
\captionof{table}{\small Results of the \emph{Table as a Paragraph} strategy on \datasetName subsets with RoBERTa$_{L}$ model, hypothesis-only baseline and majority human agreement. The first three rows are reproduced from~\citet{gupta-etal-2020-infotabs}. The last row represents the average performances (and standard deviations as subscripts) using models obtained via five-fold cross validation (5xCV).}
\label{tab:pararesults}
\end{minipage}

\section{Reasoning: An illusion?}
\label{sec:reasoning}

\emph{Given the surprisingly high accuracies in Table~\ref{tab:pararesults}, especially on the \alphaOne test dataset, can we conclude that the RoBERTa-based model reasons effectively about the evidence in the tabular input to make its inference?} That is, does it arrive at its answer via a sound logical process that takes into account all available evidence along with common sense knowledge? Merely achieving high accuracy is not sufficient evidence of reasoning: the model may arrive at the right answer for the wrong reasons leading to improper and inadequate generalization over unseen data. This observation is in line with the recent work pointing out that the high-capacity models we use may be relying on spurious correlations~\cite[e.g.,][]{poliak2018hypothesis}.

``Reasoning'' is a multi-faceted phenomenon, and fully characterizing it is beyond the scope of this work. However, we can probe for the {\em absence} of evidence-grounded reasoning via  model responses to carefully constructed inputs and their variants. The guiding premise for this work is:

\begin{quote}
\emph{Any ``evidence-based reasoning'' system should demonstrate expected, predictable behavior in response to controlled changes to its inputs.}
\end{quote}

In other words, ``reasoning failures'' can be identified by checking if a model deviates from expected behavior in response to controlled changes to inputs. We note that this strategy has been either explicitly or implicitly employed in several lines of recent work~\cite{ribeiro2020beyond,gardner2020evaluating}.
In this work, we instantiate the above strategy along three specific dimensions, briefly introduced here using the running example in Figure~\ref{fig:example}. Each dimension is used to define several concrete probes that subsequent sections detail.

\paragraph{1. Avoiding Annotation Artifacts} A model should not rely on spurious lexical correlations. In general, it should not be able to infer the label using only the hypothesis. Lexical differences in closely related hypotheses should produce predictable changes in the inferred label. For example, in the hypothesis H2 of Figure \ref{fig:example} if the token ``end'' is replaced with ``start'', the model prediction should change from \contradict to \entail.

\paragraph{2. Evidence Selection} A model should use the correct evidence in the premise for determining the hypothesis label. For example, ascertaining that the hypothesis H1 is entailed requires the \rowKey{Genre} and \rowKey{Length} rows of Figure \ref{fig:example}. When a relevant row is removed from a table, a model that predicts the \entail or the \contradict label should predict the \neutral label. When an irrelevant row is removed, it should not change its prediction from \entail to \neutral or vice versa.

\paragraph{3. Robustness to Counterfactual Changes} A model's prediction should be \emph{grounded} in the provided information even if it contradicts the real world, i.e., to counterfactual information. For example, if the month of the \rowKey{Released} date changed to ``December'', then the model should change the label of H2 in Figure \ref{fig:example} to \entail from \contradict. Since this information about release date contradicts the real world, the model cannot rely on its pre-trained knowledge, say from Wikipedia. For the model to predict the label correctly, it needs to reason with the information in the table as the primary evidence. Although the importance of pre-trained knowledge cannot be overlooked, it must not be at the expense of primary evidence.

Further, there are certain pieces of information in the premise (irrelevant to the hypothesis) which do not impact the outcome, making the outcome \emph{invariant} to these changes. For example, deleting irrelevant rows from the premise should not change the model's predicted label. Contrary to this is the relevant information (``evidence'') in the premise. Changing these pieces of information should vary the outcome in a predictable manner, making the model \emph{covariant} with these changes. For example, deleting relevant evidence rows should change the model's predicted label to \neutral.

The three dimensions above are not limited to tabular inference. They can be extended to other NLP tasks, such as reading comprehension as well as the standard sentential NLI. However, directly checking for such properties there would require a lot of labeled data---a big practical impediment. Fortunately, in the case of tabular inference, the (in-/co-)variants associated with these dimensions allow controlled and semi-automatic edits to the inputs leading to predictable variation of the expected output. This insight underlies the design of probes using which we examine the robustness of the reasoning employed by a model performing tabular inference. As we will see in the following sections, highly effective and precise probes can be designed without extensive annotation.
\section{Probing Annotation Artifacts}
\label{sec:probe2}

\emph{Can a model make inference about a hypothesis without a premise?} It is natural to answer in the negative in general (Of course, certain hypotheses may admit strong priors, e.g., tautologies.). Preliminary experiments by \citet{gupta-etal-2020-infotabs} on \datasetName, however, reveal that a model trained just on hypotheses performs surprisingly well on the test data. This phenomenon, an inductive bias entirely predicated on the hypotheses, is called \emph{hypothesis bias}. Models for other NLI tasks have been similarly shown to exhibit hypothesis bias, whereby the models learn to rely on spurious correlations between patterns in the hypotheses and corresponding labels ~\cite[][and others]{poliak2018hypothesis, gururangan2018annotation, geva2019we}. For example, negations are observed to be highly correlated with contradictions~\cite{niven-kao-2019-probing}.

To better characterize a model's reliance on such artifacts, we perform controlled edits to hypotheses without altering associated premises. Unlike the \alphaTwo set, which includes minor changes to function words, we aim to create more sophisticated changes by altering content expressions or noun phrases in a hypothesis. Two possible scenarios arise where a hypothesis alteration, without a change in the premise, either (a) leads to a change in the label (i.e., the label covaries with the variation in the hypothesis), or (b) does not induce a label change (i.e., the label is invariant to the variation in the hypothesis). 

In \datasetName, a set of reasoning categories are identified to characterize the relationship between a tabular premise and a hypothesis. We use a subset of these, listed below, to perform controlled changes in the hypotheses.
\vspace*{0.5em}
\begin{itemize} [nosep]
\setlength\itemsep{0em}
\setlength\itemindent{-10pt} 
\item {\bf Named Entities}:~such as \emph{Person, Location, Organisation}; 
\item {\bf Nominal modifiers}:~nominal phrases or clauses; 
\item {\bf Negation}:~ markers such as \emph{no, not}; 
\item {\bf Numerical Values}:~numeric expressions representing \emph{weights, percentages, areas};
\item {\bf Temporal Values}:~Date and Time; and,
\item {\bf Quantifiers}:~like \emph{most, many, every}.
\end{itemize}
\vspace*{0.5em}
Although we can easily track these expressions in a hypothesis using tools like entity recognizers and parsers, it is non-trivial to automatically modify them with a predictable change on the hypothesis label. For example, some label changes can only be controlled if the target expression in the hypothesis is correctly aligned with the facts in the premise. Such cases include \contradict to \entail, and \neutral to \contradict or \entail, which are difficult without extensive expression-level annotations. Nonetheless, in several cases, label changes can be deterministically known even with imprecise changes in the hypothesis. For example, we can convert a hypothesis from \entail to \contradict by replacing a named entity in the hypothesis with a random entity of the same type.

Hence we follow the following strategy: (a) We avoid perturbations involving the \neutral label altogether, as they often need changes in the premise (table) as well. (b) We generate all label-preserving and some label-flipping transformations automatically using the approach described below. (c) We annotate the \contradict to \entail label-flipping perturbations manually.

\paragraph{Automatic generation of label-preserving transformations} To automatically perturb hypotheses, we leverage the syntactic structure of a hypothesis and the monotonicity properties of function words like prepositions. First, we perform syntactic analysis of a hypothesis to identify named entities and their relations to title expressions via dependency paths.\footnote{We used \href{https://spacy.io/}{spaCy v2.3.2} for the syntactic analysis.} Then, based on the entity type, we either substitute or modify them. Named entities such as person names and locations are substituted with entities of the same type. Expressions containing numbers are modified using the monotonicity property of the prepositions (or other function words) governing them in their corresponding syntactic trees. 

\begin{minipage}{0.97\columnwidth}
\bigskip
  {\footnotesize
    \begin{tabular}{p{0.2\linewidth}p{0.25\linewidth}p{0.3\linewidth}}
      \toprule
     {\bf Preposition}   & {\bf Upward Monotonicity} & {\bf Downward Monotonicity}  \\ 
      \midrule
       over        &  \contradict              & \entail	 \\
       under       &  \entail                  & \contradict \\ 
       more than   &  \contradict              & \entail \\
       less than   &  \entail                  & \contradict \\
       before      &  \entail                  & \contradict \\
       after       &  \contradict              & \entail \\
      \bottomrule
    \end{tabular}}
    \captionof{table}[fontsize=\footnotesize]{\label{tab:monotonicity} \small Monotonicity properties of prepositions.}
    \bigskip
\end{minipage}

Given the monotonicity property of a preposition (see Table \ref{tab:monotonicity}), we modify its governing numerical expression in a hypothesis in the same order to preserve the hypothesis label. Consider the hypothesis H5 in Figure \ref{tbl:bridesmaids} which contains a preposition (\emph{over}) with upward monotonicity. Because of upward monotonicity, we can increase the number of hours in H5 without altering the label.

\begin{table}[!htbp]
  {\footnotesize
    \begin{tabular}{>{\raggedright}p{0.25\linewidth}p{0.6\linewidth}}
      \toprule
      {\bf Type of Modification} & {\bf Perturbed Hypothesis} \\                        
      \midrule
      {\bf Nominal Modifier	} & H1$^E_E$: Breakfast in America \textcolor{red}{\textit{which was produced by Pert Handerson}} is a pop album of 46 minutes length.	 \\
      {\bf Temporal Expression} &   H1$^E_C$: Breakfast in America is a pop album with a length of \textcolor{red}{\textit{56}} minutes. \\ 
      {\bf Negation} & H2$^C_E$: Breakfast in America was \textcolor{red}{\textit{not}} released towards the end of 1979. \\
      {\bf Temporal Expression} & H2$^C_C$: Breakfast in America was released towards the end  of \textcolor{red}{\textit{1989}}. \\
      \bottomrule
    \end{tabular}}
    \caption{\label{tab:hypo_only_pert} \small Example hypothesis perturbations for the running example from Figure~\ref{fig:example}. The \textcolor{red}{\textit{red italicized}} text represents changes. Superscripts $E/C$ represent gold \entail and \contradict labels, while subscripts $E/C$ represent new labels.}
\end{table}

\noindent {\bf Manual annotation of label-flipping transformations} Note that in the above example, modifying the numerical expression in the reverse direction (e.g., decreasing the number of hours) does not guarantee a label flip. We need to know the premise to be accurate. During the experiments, we observed that a large step (half/twice the actual number) suffices in most cases. We used this heuristic and manually curated the erroneous cases. Additionally, all the cases of \contradict to \entail label-flipping perturbations were annotated manually.\footnote{Annotation done by an expert well versed in the NLI task.}

\noindent\begin{minipage}{\columnwidth}
\bigskip
  {\footnotesize
  \begin{center}
    \begin{tabular}{>{\raggedright}p{0.25\linewidth}p{0.6\linewidth}}
      \toprule
      \multicolumn{2}{c}{\bf Bridesmaids}  \\  
      \midrule
      {\bf Length} & 	125 minutes \\
      \bottomrule
    \end{tabular}
  \end{center}
}
  {\small
    \begin{enumerate}[nosep]
    \item[H5:] Bridesmaids has a running time of over 3 hrs.
    \end{enumerate}}
    \captionof{figure}{\small \label{tbl:bridesmaids} Hypothesis H5 contradicts the premise.}
  \label{fig:hypo_contradict}
  \bigskip
\end{minipage}

We generated 2,891 perturbed examples from the \alphaOne set with 1,203 instances preserving the label and 1,688 instances flipping it. We also generated 11,550 examples from the \trainSet set, with 4,275 preserving and 7,275 flipping the label. Some example perturbations using different types of expressions are listed in Table \ref{tab:hypo_only_pert}. It should be noted that there may not be a one-to-one correspondence between the gold and perturbed examples, as a hypothesis may be perturbed numerous times or not at all. As a result, in order for the results to be comparable, a single perturbed example must be sampled for each gold example: we sampled $967$ from the \alphaOne set and $4,274$ from the \trainSet set.

\paragraph{Results and Analysis}
We tested the hypothesis-only and full models (both trained on the original \trainSet set) on the perturbed examples, without subsequent fine-tuning on the perturbed examples.\footnote{We analyse the impact of  fine-tuning on perturbed examples in \S\ref{sec:inoculation}.} The results are presented in Table \ref{tab:hypothesis_only_results}, with each cell representing the average accuracy and standard deviation (subscript) across 100 samplings, with 80$\%$ of the data selected at random in each sampling.

We note that the performance degrades substantially in both label-preserved and flipped settings when the model is trained on just the hypotheses. When labels are flipped after perturbations, the decrease in performance (averaged across both models) is about $25\%$ and $61\%$ points, on the \alphaOne set and \trainSet set respectively. However, for the full model, perturbations that retain the hypothesis label have little effect on model performance. 

The contrast in the performance drop between the label-preserved and label-flipped cases suggests that changes to the content expressions have little effect on the model's original predictions. Interestingly, the predictions are invariant to changes to functions words as well, as per results on \alphaTwo in \citet{gupta-etal-2020-infotabs}. This suggests that the model might be more prone to changes to the template or structure of a hypothesis than its lexical makeup. Consequently, a model that relies on correlations between the hypothesis structure and the label is expected to suffer on the label-flipped cases. In case of label-preserving perturbations of similar kind, structural correlations between the hypothesis and the label are retained leading to minimal drop in model performance.

\noindent\begin{minipage}{0.95\columnwidth}
\bigskip
     {\footnotesize
     \centering
     \begin{tabular}{p{1.55cm}|p{1.5cm}|p{1.5cm}|p{1.2cm}}
     \toprule
     {\bf Model} & {\bf Original} ${mean}_{({stdev})}$ & {\bf Label Preserved}  &  {\bf Label Flipped}  \\ \midrule
    \multicolumn{4}{c}{\bf \trainSet Set (w/o \neutral)} \\ \midrule
     Prem+Hypo & 99.44$_{(0.06)}$ & 92.98$_{(0.20)}$  & 53.92$_{(0.28)}$ \\
     Hypo-Only & 96.39$_{(0.13)}$ & 70.23$_{(0.35)}$  & 19.23$_{(0.27)}$\\
     \midrule
      \multicolumn{4}{c}{\bf \alphaOne Set (w/o \neutral)} \\ \midrule
     Prem+Hypo & 68.94$_{(0.76)}$ & 69.56$_{(0.77)}$  & 51.48$_{(0.86)}$ \\
     Hypo-Only & 63.52$_{(0.75)}$ & 60.27$_{(0.85)}$  & 31.02$_{(0.63)}$ \\
     \bottomrule     
     \end{tabular}}
     \captionof{table}{\small Results of the Hypothesis-only model and Prem+Hypo model on the gold and perturbed hypotheses.}
    \label{tab:hypothesis_only_results}
    \bigskip
 \end{minipage}
 
The results of the hypothesis-only model on the \trainSet set may appear slightly surprising at first. However, given that the model was trained on this dataset, it seems reasonable to assume that the model has `overfit` to the training data. Therefore, the model is expected to be vulnerable even to slight label-preserving modifications to the examples it was trained on, leading to the huge drop of 26$\%$. In the same setting, for the \alphaOne set the performance drop is lesser, namely about $3\%$. 

Taken together, we can conclude from these results that the model ignores the information in the hypotheses, (thereby perhaps also the aligned facts in the premise), and instead relies on irrelevant structural patterns in the hypotheses.
\section{Probing Evidence Selection }
\label{sec:probe1}
Predictions of an NLI model should primarily be based on the evidence in the premise, that is, on the facts relevant to the hypothesis. For a tabular premise, rows containing the evidence necessary to infer the associated hypothesis are called relevant rows. Short-circuiting the evidence in relevant rows for inference using annotation artifacts as suggested in \S\ref{sec:probe2} or other spurious artifacts in irrelevant rows of the table is expected to lead to poor generalization over unseen data. 

To better understand the model's ability to select evidence in the premise, we use two kinds of controlled edits: (a) automatic edits without any information about relevant rows, and, (b) semi-automatic edits using knowledge of relevant rows via manual annotation. The rest of the section goes over both scenarios in detail. All experiments in this section use the full model that is trained on both premises and their associated hypotheses.

\subsection{Automatic Probing}
\label{sec:probe1automatic}
We define four kinds of table modifications that are agnostic to the relevance of rows to a hypothesis: \begin{inparaenum}[(a)] 
 \item \textit{row deletion}, 
 \item \textit{row insertion}, 
 \item \textit{row-value update}, i.e., changing existing information, and 
 \item \textit{row permutation}, i.e., reordering rows.
 \end{inparaenum}
Each modification allows certain desired (valid) changes to model predictions.\footnote{In performing these modifications, we made sure that the modified table does not become inconsistent or self-contradicting.} We examine below the case of row deletion in detail and  refer the reader to the \hyperref[app:appendix-automatic-probing]{Appendix} for the others.

{\noindent \bf Row deletion} should lead to the following desired effects: (a) If the deleted row is relevant to the hypothesis (e.g., \rowKey{Length} for H1), the model prediction should change to \neutral. (b) If the deleted row is irrelevant (e.g., \rowKey{Producer} for H1), the model should retain its original prediction. \neutral predictions should remain unaffected by row deletion.

\begin{minipage}{.94\columnwidth}
\bigskip
\centering
  {\small \begin{tikzpicture}[thick, scale=0.6, {every node/.style}={scale=0.6}]
\begin{pgfonlayer}{nodelayer}
		\node [style=new style 0] (0) at (0, 8) {C};
		\node [style=new style 0] (1) at (-7, 0) {E};
		\node [style=new style 0] (2) at (7, 0) {N};
		\node [style=none] (3) at (-6.5, 1.5) {};
		\node [style=none] (4) at (-6, 1) {};
		\node [style=none] (5) at (-5.75, -0.5) {};
		\node [style=none] (6) at (-5.75, 0.25) {};
		\node [style=none] (7) at (-1, 7.25) {};
		\node [style=none] (8) at (-0.5, 6.75) {};
		\node [style=none] (9) at (0.75, 6.75) {};
		\node [style=none] (10) at (1.25, 7.25) {};
		\node [style=none] (11) at (6, 1) {};
		\node [style=none] (12) at (6.5, 1.5) {};
		\node [style=none] (13) at (5.5, -0.5) {};
		\node [style=none] (14) at (5.5, 0.25) {};
		\node [style=new style 1] (20) at (0, 11.5) {90.75, 91.44, 91.08};
		\node [style=new style 1] (21) at (4.25, 7.75) {6.02, 4.86, 5.91};
		\node [style=new style 1] (22) at (7.5, -3.25) {95.57, 96.1, 94.76};
		\node [style=new style 1] (23) at (0, -1.25) {6.03, 6.64, 4.98};
		\node [style=new style 1] (24) at (-7.25, -3.25) {88.21, 86.10, 90.01};
		\node [style=new style 2] (25) at (6.75, 4.25) {2.67, 2.36, 2.95};
		\node [style=new style 2] (29) at (-7, 4) {5.76, 7.26, 5.01};
		\node [style=new style 2] (30) at (-5, 8) {3.23, 3.7, 3.01};
		\node [style=new style 2] (33) at (0, 1) {1.76, 1.55, 2.29};
		\node [style=none] (34) at (-5, 7.5) {};
		\node [style=none] (35) at (-2.5, 4.75) {};
		\node [style=none] (36) at (3, 4.25) {};
		\node [style=none] (37) at (4.25, 7.25) {};
		\node [style=none] (38) at (0.75, 6.25) {\bf *};
		\node [style=none] (39) at (7.75, -2) {*};
	\end{pgfonlayer}
	\begin{pgfonlayer}{edgelayer}
		\draw [style=new edge style 0] (3.center) to (7.center);
		\draw [style=new edge style 0] (8.center) to (4.center);
		\draw [style=new edge style 0] (14.center) to (6.center);
		\draw [style=new edge style 0] (12.center) to (10.center);
		\draw [style=new edge style 3, in=-120, out=-45, loop] (2) to ();
		\draw [style=new edge style 5, in=-135, out=-45, loop] (1) to ();
		\draw [style=new edge style 5, in=135, out=45, loop] (0) to ();
		\draw [style=new edge style 4] (5.center) to (13.center);
		\draw [style=new edge style 4] (9.center) to (11.center);
		\draw [style=new edge style 5, bend left=45, looseness=1.75] (35.center) to (34.center);
		\draw [style=new edge style 5, bend right=45, looseness=1.25] (36.center) to (37.center);
	\end{pgfonlayer}
\end{tikzpicture}}
    
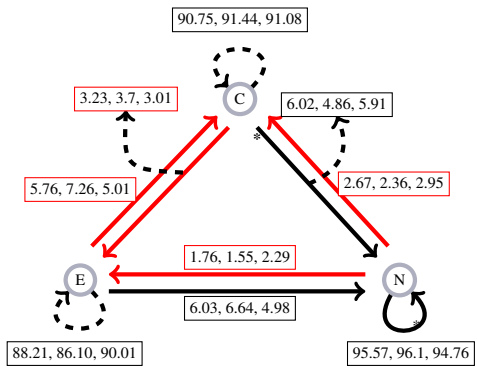
\captionof{figure}{\small Changes in model predictions after automatic row deletion. Directed edges are labeled with transition percentages from the source node label to the target node label. The number triple corresponds to $\alpha_1$, $\alpha_2$ and $\alpha_3$ test sets respectively and for each source node, adds up to 100$\%$ over the outgoing edges. \textcolor{red}{Red} lines represent invalid transitions. Dashed and solid \textbf{black} lines represent valid transitions for irrelevant and relevant row deletion respectively. * represents valid transitions with either row deletions.}
    \label{fig:probe1deletion}
\end{minipage}

\paragraph{Results and Analysis} We studied the impact of \textit{row deletion} on the \alphaOne, \alphaTwo and \alphaThree test sets. Figure~\ref{fig:probe1deletion} shows aggregate changes to labels after row deletions as a directed labeled graph. The nodes in this graph represent the three labels in \datasetName, and the edges denote transitions after {\it row deletion}. The source and end nodes of an edge represent predictions before and after the modification.

We see that the model makes invalid transitions in all three datasets. Table \ref{tab:summary_rowdeletion} summarizes the invalid transitions by aggregating them over the label originally predicted by the model. The percentage of invalid transitions is higher for \entail predictions than for \contradict and \neutral. After row deletion, many \entail examples are incorrectly transitioning to \contradict rather than to \neutral. The opposite trend is observed for the \contradict predictions.

\noindent\begin{minipage}[!t]{.95\columnwidth}
\bigskip
{\small
    \begin{tabular}{l|c|c|c|c}
    \toprule
    Dataset & \alphaOne & \alphaTwo & \alphaThree & Average\\  \midrule
        \entail & 5.76 & 7.26 & 5.01 & 6.01 \\
        \neutral & 4.43 & 3.91 & 5.24 & 4.53\\
        \contradict & 3.23 & 3.70 & 3.01 & 3.31\\ \midrule
        Average & 4.47 & 4.96  & 4.42 & - \\
        \bottomrule
    \end{tabular}}
    \captionof{table}{\small Percentage of invalid transitions after row deletion. For an ideal model, all these numbers should be zero.}
    \label{tab:summary_rowdeletion}
 \bigskip
\end{minipage}

As with row deletion, the model exhibits invalid responses to other row modifications listed in the beginning of this section, like row insertion. Surprisingly, the performance degrades due to row permutations as well, suggesting some form of position bias in the model. On the positive side, the model mostly retains its predictions on row-value update operations. We refer the reader to the  \hyperref[app:appendix-automatic-probing]{Appendix} for more details.

\subsection{Manual Probing}
\label{sec:probe1manual}
Row modification for automatic probing in \S\ref{sec:probe1automatic} is agnostic to the relevance of the row to a given hypothesis. Since only a few rows (one or two) are relevant to the hypothesis, the probing skew towards hypothesis-unrelated rows weakens the investigations into the evidence-grounding capability of the model. 
Knowing the relevance of rows allows for the creation of stronger probes. For example, if a relevant row is deleted, the \entail and \contradict predictions should change to \neutral. (Recall that after deleting an irrelevant row the model should retain its original label.)

Probing by altering or deleting relevant rows requires human annotation of relevant rows for each table-hypothesis pair. We used MTurk to annotate the relevance of rows in the development and the test sets, with turkers identifying the relevant rows for each table-hypothesis pair.

\paragraph{Inter-annotator Agreement.} 
We employed majority voting to derive ground truth labels from multiple annotations for each row. The inter-annotator agreement macro F1 score for each of the four datasets is over $90\%$ and the average Fleiss' kappa is $78$ (std: $0.22$). This suggests good inter-annotator agreement. In 82.4$\%$ of cases, at least $3$ out of $5$ annotators marked the same relevant rows.

\paragraph{Results and Analysis} We examined the response of the model when relevant rows are deleted. Figure \ref{fig:probe1deletionrelevant} shows the label transitions. The fact that even after the deletion of relevant rows, \entail and \contradict predictions don't change to \neutral a large percentage of times (mostly the original label remains unchanged and at other times, it changes incorrectly), indicates that the model is likely utilizing spurious statistical patterns in the data for making the prediction. 

\begin{minipage}[b]{0.93\columnwidth}
\bigskip
\centering
    \begin{tikzpicture}[thick, scale=0.6, {every node/.style}={scale=0.6}]
	\begin{pgfonlayer}{nodelayer}
		\node [style=new style 0] (0) at (0, 8) {C};
		\node [style=new style 0] (1) at (-7, 0) {E};
		\node [style=new style 0] (2) at (7, 0) {N};
		\node [style=none] (3) at (-6.5, 1.5) {};
		\node [style=none] (4) at (-6, 1) {};
		\node [style=none] (5) at (-5.75, -0.5) {};
		\node [style=none] (6) at (-5.75, 0.25) {};
		\node [style=none] (7) at (-1.25, 7.25) {};
		\node [style=none] (8) at (-0.75, 6.75) {};
		\node [style=none] (9) at (0.75, 6.75) {};
		\node [style=none] (10) at (1.25, 7.25) {};
		\node [style=none] (11) at (6, 1) {};
		\node [style=none] (12) at (6.5, 1.5) {};
		\node [style=none] (13) at (5.5, -0.5) {};
		\node [style=none] (14) at (5.5, 0.25) {};
		\node [style=new style 1] (20) at (0, 11.75) {};
		\node [style=new style 1] (22) at (7.5, -3.75) {91.62, 93.42, 91.99};
		\node [style=new style 1] (23) at (0, 1) {};
		\node [style=new style 1] (24) at (-7.25, -3.75) {};
		\node [style=new style 2, rotate=47] (29) at (-4.75, 4.5) {21.00, 23.42, 16.64};
		\node [style=new style 2, rotate=47] (30) at (-3, 3.25) {4.89, 7.67, 3.53};
		\node [style=new style 2] (33) at (0, -1.25) {};
		\node [style=new style 2] (48) at (0, 1) {1.07, 1.05, 2.04};
		\node [style=new style 1] (49) at (0, -1.25) {};
		\node [style=new style 1] (50) at (0, -1.25) {24.59, 25.3, 22.68};
		\node [style=new style 2] (51) at (0, 11.75) {};
		\node [style=new style 2] (52) at (0, 11.75) {72.13, 73.43, 74.27};
		\node [style=new style 2] (53) at (-7.25, -3.75) {54.41, 51.28, 60.67};
		\node [style=new style 2, rotate=-47] (56) at (4.25, 5) {7.32, 5.53, 5.97};
		\node [style=new style 1, rotate=-47] (57) at (2.75, 3.5) {22.99, 18.90, 22.21};
	\end{pgfonlayer}
	\begin{pgfonlayer}{edgelayer}
		\draw [style=new edge style 0] (8.center) to (4.center);
		\draw [style=new edge style 0] (14.center) to (6.center);
		\draw [style=new edge style 3, in=-135, out=-45, loop] (2) to ();
		\draw [style=new edge style 3] (13.center) to (5.center);
		\draw [style=new edge style 0] (3.center) to (7.center);
		\draw [style=new edge style 1] (11.center) to (9.center);
		\draw [style=new edge style 0] (12.center) to (10.center);
		\draw [style=new edge style 0, in=-135, out=-45, loop] (1) to ();
		\draw [style=new edge style 0, in=135, out=45, loop] (0) to ();
	\end{pgfonlayer}
\end{tikzpicture}
    \captionof{figure}{\small Changes in model predictions after deletion of relevant rows. \textcolor{red}{Red} lines represent invalid transitions while \textbf{black} lines represent valid transitions. The directed edges are labeled in the same manner as they are in Figure \ref{fig:probe1deletion}.}
    \label{fig:probe1deletionrelevant}
\bigskip
\end{minipage}

\begin{minipage}[b]{0.95\columnwidth}
    \centering
    {\small
    \begin{tabular}{l|p{0.7cm}|p{0.6cm}|p{0.6cm}|p{1cm}}
    \toprule
        Dataset & \alphaOne & \alphaTwo & \alphaThree & Average \\ \midrule
        \entail & 75.41 & 74.70 &  77.31 & 75.80\\
        \neutral & 8.39 & 6.58 & 8.01 & 7.66\\
        \contradict & 77.02 & 81.10 & 77.80 & 78.64\\ \midrule
        Average & 53.60 & 54.14 & 54.35 & \\
        \bottomrule
    \end{tabular}}
    \captionof{table}{\small Percentage of invalid transitions following deletion of relevant rows. For an ideal model, all these numbers should be zero.}
    \label{tab:summary_rowrelevant_deletion}
    \smallskip
\end{minipage}

We summarize the combined invalid transitions for each label in Table \ref{tab:summary_rowrelevant_deletion}. We see that the percentage of invalid transitions is considerably higher compared to random row deletion in Figure \ref{fig:probe1deletion}.\footnote{Note that the dashed black lines from Figure \ref{fig:probe1deletion} are now red in Figure \ref{fig:probe1deletionrelevant}, indicating invalid transitions.} The large percentage of invalid transitions in the \entail and \contradict cases indicates a rather high utilization of spurious statistical patterns by the model to arrive at its answers. 

\subsection{Human vs Model Evidence Selection}
We further analyze the model's capability for selecting relevant evidence by comparing it with human annotators. All rows that alter the model predictions during automatic row deletion are considered as \emph{model relevant rows} and are compared to the human-annotated relevant rows. We only consider the subset of $4600$ (from $7200$ annotated dev/test sets pairs) hypothesis-table pairs with \entail and \contradict labels, where deleting a relevant row should change the prediction to \neutral.\footnote{We did not include the $2400$ \neutral examples pairs and the ambiguous $200$ \entail or \contradict examples that had no relevant rows as per the consensus annotation.} 

\paragraph{Results and Analysis} On the human-annotated relevant rows, the model has an average precision of 41.0\% and a recall of 40.9\%. Further analysis reveals that the model \begin{inparaenum}[(a)] \item  uses all relevant rows in $27\%$ cases, \item uses incorrect or no rows as evidence in $52\%$ of occurrences, and \item is only partially accurate in identifying relevant rows in the remaining $21\%$ of examples. \end{inparaenum} Upon further analysing the cases in (b), we observed that the model actually ignores premises completely in 88$\%$ (of $52\%$) of cases. This accounts for 46$\%$ (absolute) of all occurrences. In comparison, in the human-annotated data, such cases only amount to $<$ 2$\%$. 

Although, the model's predictions are 70$\%$ correct in the 4,600 examples, only 21$\%$ can be attributed to using all relevant evidence. The correct label in 37$\%$ of the 4,600 examples is from irrelevant rows, with the remaining 12$\%$ of correct predictions use some, but not all, relevant rows.

We can conclude from the findings in this section that the model does not seem to need all the relevant evidence to arrive at its predictions, raising questions about trust in its predictions.

\section{Probing with Counterfactual Examples}
\label{sec:probe3}

Since \datasetName is a dataset of facts based on Wikipedia, pre-trained language models such as RoBERTa, trained on Wikipedia and other publicly available text, may have already encountered information in \datasetName during pre-training. As a result, NLI models built on top of RoBERTa$_{L}$ can learn to infer a hypothesis using the knowledge of the pre-trained language model. More specifically, the model may be relying on ``\emph{confirmation bias}'', in which it selects evidence/patterns from both premise and hypothesis that matches its prior knowledge. While world knowledge is necessary for table NLI~\cite{neeraja-etal-2021-infotabskg}, models should still treat the premise as the primary evidence. 

Counterfactual examples can help test whether the model is grounding its inference on the evidence provided in the tabular premise. In such examples, the tabular premise is modified such that the content does not reflect the real world. In this study, we limit ourselves to modifying only the \entail and \contradict examples. We omit the \neutral cases because the majority of them in \datasetName involve out-of-table information; producing counterfactuals for them is much harder and involves the laborious creation of new rows with the right information.

The task of creating counterfactual tables presents two challenges. First, the modified tables should not be self-contradictory. Second, we need to determine the labels of the associated hypotheses after the table is modified. We employ a simple approach to generate counterfactuals that addresses both challenges. We use the evidence selection data (\S\ref{sec:probe1manual}) to gather all premise-hypothesis pairs that share relevant keys such as ``Born'', ``Occupation'' etc. Counterfactual tables are generated by swapping the values of relevant keys from one table to another.\footnote{There may still be a few cases of self-contradiction, but we expect that such invalid cases would not exist in the rows that are relevant to the hypothesis.}

Figure~\ref{fig:examplecounterfactual1} shows an example. We create counterfactuals from the premises in Figure \ref{fig:example} and Figure \ref{fig:hypo_contradict} by swapping their {\bf Length} rows. We also swap the hypotheses ({H1 and H5}) aligned to the {\bf Length} rows in both premises by replacing the title expression {\bf Bridesmaids} in H5 with {\bf Breakfast in America} and \emph{vice versa}. The simple procedure ensures that the hypotheses labels are left unchanged in the process, resulting in high-quality data. 

In addition, we also generated counterfactuals by swapping the table title and associated expressions in the hypotheses with the title of another table, resulting in a counterfactual table-hypothesis pair, as in the row swapping strategy. Figure~\ref{fig:examplecounterfactual2} shows an example created from the premises in Figure \ref{fig:example} and Figure \ref{fig:hypo_contradict} by swapping the title rows {\bf Breakfast in America} and {\bf Bridesmaids}. The title expression in all hypotheses in Figure \ref{fig:example} are also replaced by {\bf Bridesmaids}. This strategy also preserves the hypothesis label similar to row swapping.

\noindent\begin{minipage}[b]{0.96\columnwidth}
\bigskip
  \centering
  {
  \footnotesize
  \begin{center}
    \begin{tabular}{>{\raggedright}p{0.25\linewidth}p{0.6\linewidth}}
      \toprule
      \multicolumn{2}{c}{\bf Breakfast in America} \\
      \midrule
      {\bf Released	} & 	29 March 1979 \\
      {\bf Recorded} & May–December 1978 \\ 
      {\bf Studio}  &  The Village Recorder (Studio B) in Los Angeles  \\  
      {\bf Genre}        & pop ; art rock ; soft rock  \\
      \boxit{\linewidth}{\bf Length}    & \boxit{\linewidth}125 minutes \\  
     {\bf Label}             & A$\&$M  \\ 
     {\bf Producer} & 	Peter Henderson, Supertramp \\
      \bottomrule
    \end{tabular}
  \end{center}
}
  {\small
    \begin{enumerate}[nosep]
    \item[$\widehat{H5}$:] \textcolor{red}{Breakfast in America} has a running time of over 3 hrs.
    \end{enumerate}}
    \captionof{figure}{\small Counterfactual table-hypothesis pair created from Figure \ref{fig:example} and Figure \ref{fig:hypo_contradict}. Only the values of `Length' rows are swapped, rest of the rows from Figure \ref{fig:example} are copied over.}
  \label{fig:examplecounterfactual1}
  \bigskip
\end{minipage}

\noindent\begin{minipage}{0.96\columnwidth}
  \centering
  {
  \footnotesize
  \begin{center}
    \begin{tabular}{>{\raggedright}p{0.25\linewidth}p{0.6\linewidth}}
      \toprule
      \multicolumn{2}{c}{\bf \textcolor{red}{Bridesmaids}} \\  
      \midrule
      {\bf Released	} & 	29 March 1979 \\
      {\bf Recorded} & May–December 1978    \\ 
      {\bf Studio}           &  The Village Recorder (Studio B) in Los Angeles \\  
      {\bf Genre}                  & pop ; art rock ; soft rock \\ 
      {\bf Length}      & 46:06 \\  
     {\bf Label}             & A$\&$M   \\ 
     {\bf Producer} & 	Peter Henderson, Supertramp\\
      \bottomrule
    \end{tabular}
  \end{center}
}
  {\footnotesize
    \begin{enumerate}[nosep]
    \item[$\widehat{H1}$:] \textcolor{red}{Bridesmaids} is a pop album with a length of 46 minutes.
    \item[$\widehat{H2}$:] \textcolor{red}{Bridesmaids} was released at the end of 1979.
    \item[$\widehat{H3}$:] Most of \textcolor{red}{Bridesmaids} was recorded in the last month of 1978.
    \item[$\widehat{H4}$:] \textcolor{red}{Bridesmaids} has 6 tracks.
    \end{enumerate}
  }
    \captionof{figure}{\small A counterfactual tabular premise and the associated hypotheses created from Figures \ref{fig:example} and \ref{fig:hypo_contradict}. The hypotheses $\widehat{H1}$ is entailed by the premise, $\widehat{H2}$ contradicts it, and $\widehat{H3}$ and $\widehat{H4}$ are neutral.}
  \label{fig:examplecounterfactual2}
  \bigskip
\end{minipage}

The above approaches are \emph{Label Preserving} as they do not alter the entailment labels. Counterfactual pairs with flipped labels are important for filtering out the contribution of artifacts or other spurious correlations that originate from a hypothesis (see \S\ref{sec:probe2}). So, in addition, we also created counterfactual table-hypothesis pairs where the original labels are flipped. These counterfactual cases are, however, non-trivial to generate automatically, and are therefore created manually. To create the \emph{Label-Flipped} counterfactual data, three annotators manually modified tables from the \trainSet and \alphaOne datasets corresponding to \entail and \contradict labels, producing $885$ counterfactual examples from the \trainSet set and $942$ from the \alphaOne set. The annotators cross-checked the labels to determine annotation accuracy, which was 88.45$\%$ for the \trainSet set and 86.57$\%$ for the \alphaOne set.

\paragraph{Results and Analysis} We tested both hypothesis-only and full (Prem+Hypo) models on the counterfactual examples created above, without fine-tuning on a subset of these examples. The results are presented in Table \ref{tab:counterfactual_results} where each cell represents average accuracy and standard deviation (subscript) over 100 sets of 80$\%$ randomly sampled counterfactual examples. We see that the (Prem+Hypo) model is not robust to counterfactual perturbations. On the label-flipped counterfactuals, the performance drops down to close to a random prediction (48.70$\%$ for \trainSet and 44.01$\%$ for \alphaOne). The performance on the label-preserved counterfactuals is relatively better which leads us to  conjecture that the model largely exploits artifacts in hypotheses. 

\noindent\begin{minipage}{.93\columnwidth}
    \bigskip
     {\footnotesize
     \begin{tabular}{p{1.55cm}|p{1.5cm}|p{1.4cm}|p{1.2cm}}
     \toprule
     {\bf Model} & {\bf Original} ${mean}_{({stdev})}$ & {\bf Label Preserved}  &  {\bf Label Flipped}  \\ \midrule
    \multicolumn{4}{c}{\bf \trainSet Set (without \neutral)} \\ \midrule
     Prem+Hypo & 94.38$_{(0.39)}$ & 78.53$_{(0.65)}$  & 48.70$_{(0.72)}$ \\
     Hypo-Only & 99.94$_{(0.06)}$ & 82.23$_{(0.65)}$  & 00.06$_{(0.01)}$\\
     \midrule
      \multicolumn{4}{c}{\bf \alphaOne Set (without \neutral)} \\ \midrule
     Prem+Hypo & 71.99$_{(0.69)}$ & 69.65$_{(0.78)}$  & 44.01$_{(0.72)}$ \\
     Hypo-Only & 60.89$_{(0.76)}$ & 58.19$_{(0.91)}$  & 27.68$_{(0.65)}$ \\
     \bottomrule     
     \end{tabular}}
    \captionof{table}{\small Results of the Hypothesis-only and Prem+Hypo models on the gold and counterfactual examples.}
    \label{tab:counterfactual_results}
    \bigskip
\end{minipage}
 
Due to over-fitting, the \trainSet set has a larger drop of 15.85\%, compared to only 2.70\% on the \alphaOne set on the label-preserved examples. Moreover, the drop in performance for both {\it Prem+Hypo} and {\it Hypo-Only} models is comparable to their performance drop on the original table-hypothesis pairs. This shows that, regardless of whether the relevant information in the premise is accurate, both models rely substantially on hypothesis artifacts. On the \emph{Label-Flipped} counterfactuals, the large drop in accuracy could be due to both ambiguous hypothesis artifacts or counterfactual information.

To disentangle these two factors, we can take advantage of the fact that the counterfactual examples are constructed from, and hence paired with, the original examples. This allows us to examine pairs of examples where the full model makes an incorrect prediction on one, but not the other. Especially of interest are the cases where the full model makes a correct prediction on the original example, but not on the corresponding counterfactual example. 

\noindent\begin{minipage}{0.95\columnwidth}
    {\footnotesize
    \begin{tabular}{p{1cm}|p{1cm}|p{1.5cm}|p{1cm}|p{0.55cm}}
    \toprule
    \multicolumn{2}{c|}{Prem+Hypo} & Hypo-Only & \multicolumn{2}{c}{Dataset} \\
    \midrule
    C-THP &	O-THP	 & O-Hypo & \trainSet & \alphaOne\\ \midrule
    \cmark &	\xmark & \xmark & 0.00 &11.43\\ 
     \xmark &	\cmark & \xmark & 0.00 & 11.79\\
    \cmark & 	\xmark & \cmark & 3.57 & 6.48\\ 
     \xmark &	\cmark & \cmark & 49.36 & 33.12\\ 
    \bottomrule
    \end{tabular}}
    \captionof{table}{\small Performance of the full and hypothesis-only models on the original and counterfactual examples. O-THP and C-THP represent original and counterfactual table-hypothesis pairs; O-Hypo represents hypotheses from the original data; \cmark~ represents correct predictions and \xmark~ represents incorrect predictions.}
    \label{tab:model_prediction_pairing_withhypo}
\bigskip
\end{minipage}

Table \ref{tab:model_prediction_pairing_withhypo} shows the results of this analysis. Each row represents a condition corresponding to whether the full and the hypothesis-only models are correct  on the original example. The two cases of interest, described above, correspond to the second and fourth rows of the table.
The second row shows the case where the full model is correct on the original example (and not on the counter-factual example), but the hypothesis-only model is not. Since we can discount the impact of hypothesis bias in these examples, the error in the counter-factual version could be attributed to reliance on pre-trained knowledge. Unsurprisingly, there are no such examples in the training set. In the \alphaOne set, we see a substantial fraction of counterfactual examples (11.79\%) belong to this category.
The last row considers the case where the hypothesis-only model is correct. We see that this accounts for a larger fraction of the counterfactual errors, both in the training and the \alphaOne sets. Among these examples, \emph{despite} the (albeit unfortunate) fact that the hypothesis alone can support a correct prediction, the model's reliance on its pre-trained knowledge leads to errors in the counterfactual cases.

The results, taken in aggregate, suggest that the model produces predictions based on hypothesis artifacts and pre-trained knowledge rather than the evidence presented to it, thus impacting its robustness and generalization.

\noindent\begin{minipage}{0.97\columnwidth}
\bigskip
\small
     \begin{tabular}{p{1.7cm}|p{1.4cm}|p{1.2cm}|p{1.4cm}}
     \toprule
     {\bf $\#$Samples} & {\bf \alphaOne} & {\bf \alphaTwo}  &  {\bf \alphaThree}  \\ \midrule
     0 (w/o Ino) & \bf74.88 &\bf 65.55 &\bf 64.94 \\ \hdashline
     100  & 67.44 & 62.17  & 58.51 \\
     200  & 67.34 & 61.88 & 58.61 \\
     300 & 67.24 & 61.84 & 58.62 \\
     \bottomrule     
     \end{tabular}
     \captionof{table}{\small Performance of the inoculated models on the original \datasetName test sets.}
    \label{tab:hypothesis_only_results_innoculation_original}
\end{minipage}

\noindent\begin{minipage}{0.97\columnwidth}
\bigskip
     {\small
     \begin{tabular}{p{1.6cm}|p{1.4cm}|p{1.5cm}|p{1.3cm}}
     \toprule
     {\bf $\#$Samples} & {\bf Original} & {\bf Label Preserved}  &  {\bf Label Flipped}  \\ \midrule
    \multicolumn{4}{c}{\bf \trainSet Set (w/o \neutral)} \\ \midrule
     0 (w/o Ino) &\bf 99.44 & 92.98  & 53.92 \\ \hdashline
     100  & 97.24 & 95.58 &\bf 79.25\\
     200  & 97.24 &\bf 95.65 & 78.75\\
     300  & 97.24 & 95.64 & 78.74\\

     \midrule
      \multicolumn{4}{c}{\bf \alphaOne Set (w/o \neutral)} \\ \midrule
     0 (w/o Ino) &\bf 68.94 &\bf 69.56  & 51.48\\ \hdashline
     100  & 68.05 & 65.67 &\bf 57.91 \\
     200  & 68.37 & 66.29 & 57.49 \\
     300  & 68.36 & 66.29 & 57.49\\
     \bottomrule     
     \end{tabular}}
     \captionof{table}{\small Performance of the inoculated models on the hypothesis perturbed \datasetName sets.}
    \label{tab:hypothesis_only_results_innoculation_adversarial}
\end{minipage}

\section{Inoculation by Fine-Tuning}
\label{sec:inoculation}
Our probing experiments demonstrate that the models, trained on the \datasetName training set, failed along all three dimensions that we investigated. This leads us to the following question: \emph{Can additional fine-tuning with perturbed examples help?}

\citet{liu2019inoculation} point out that poor performance on challenging datasets can be ascribed to either a weakness in the model, a lack of diversity in the dataset used for training or information leakage in the form of artifacts.\footnote{Model weakness is the inherent inability of a model (or a model family) to handle certain linguistic phenomena.} They suggest that models can be further fine-tuned on a few challenging examples to determine the possible source of degradation. Inoculation can lead to one of three outcomes: 
\begin{inparaenum}[(a)]
\item {\bf Outcome 1:} The performance gap between the challenge and the original test sets reduces, possibly due to addition of diverse examples,
\item {\bf Outcome 2:} Performance on both the test sets remains unchanged, possibly because of the model's inability to adapt to the new phenomena or the changed data distribution, or,
\item {\bf Outcome 3:} Performance degrades on the test set, but improves on the challenge set, suggesting that adding new examples introduces ambiguity or contradictions.
\end{inparaenum} 

We conducted two sets of inoculation experiments to help categorize performance degradation of our models into one of these three categories. For each experiment described below, we generated additional inoculation datasets with 100, 200 and 300 examples to inoculate the original task-specific RoBERTa$_{L}$ models trained on both premises and hypotheses. As in the original inoculation work, we created these adversarial datasets by sub-sampling inclusively, i.e., the smaller datasets are subsets of the larger ones. Following the training protocol in \citet{liu2019inoculation}, we tried learning rates of $10^{-4}$, $5\times 10^{-5}$ and $10^{-5}$. We performed inoculation  for a maximum of 30 epochs with early stopping based on the development set accuracy. We found that with the first two learning rates, the model does not converge, and underperforms on the development set. The model performance was best with the learning rate of ${10^{-5}}$, which we used throughout the inoculation experiments. The standard deviation over 100 sample splits for all experiments was $\leq$ 0.91.

\paragraph{Annotation Artifacts}
Table \ref{tab:hypothesis_only_results_innoculation_original} shows the performance of the inoculated models on the original \datasetName test sets, and Table \ref{tab:hypothesis_only_results_innoculation_adversarial} shows the results on the hypothesis-perturbed examples (from \S\ref{sec:probe2}). We see that fine-tuning on the hypothesis-perturbed examples decreases performance on the original \alphaOne, \alphaTwo and \alphaThree test sets, but performance improves on the more difficult label-flipped examples of the hypothesis-perturbed test set.

\paragraph{Counterfactual Examples}
Tables \ref{tab:counterfactual_inoculation_original} and \ref{tab:counterfactual_inoculation_perturb} show the performance of models inoculated on the original \datasetName test sets and the counterfactual examples from \S\ref{sec:probe3} respectively. Once again, we see that fine-tuning on counterfactual examples improves performance on the adversarial counterfactual examples test set, at the cost of performance on the original test sets.

\paragraph{Analysis} We see that both experiments above belong to Outcome 3, where the performance improves on the challenge set, but degrades on the test set(s). The change in the distribution of inputs hurts the model: we conjecture that this may be because the RoBERTa$_{L}$ model exploits data artifacts in the original dataset but fails to do so for the challenge dataset and vice versa. 

We expect our model to handle both original and challenge datasets, at least after fine-tuning (i.e., it should belong to Outcome 1). Its failure points to the need for better models or training regimes.

\noindent\begin{minipage}{0.97\columnwidth}
\bigskip
   \small
     \begin{tabular}{p{1.7cm}|p{1.2cm}|p{1.2cm}|p{1.2cm}}
     \toprule
     {\bf $\#$Samples} & {\bf \alphaOne} & {\bf \alphaTwo}  &  {\bf \alphaThree}  \\ \midrule
     0 (w/o Ino) &\bf 74.88 &\bf 65.55 &\bf 64.94 \\ \hdashline
     100  & 69.72 &  63.88 & 59.66 \\
     200  & 69.88 & 63.78 & 58.89  \\
     300 & 67.34 & 62.23  & 57.58 \\
     \bottomrule     
     \end{tabular}
     \captionof{table}{\small Performance after inoculation by fine-tuning on the original \datasetName test sets.}
    \label{tab:counterfactual_inoculation_original}
\end{minipage}

\noindent\begin{minipage}{0.97\columnwidth}
\bigskip
     {\small
     \centering
     \begin{tabular}{p{1.6cm}|p{1.4cm}|p{1.5cm}|p{1.4cm}}
     \toprule
     {\bf $\#$Samples} & {\bf Original} & {\bf Label Preserved}  &  {\bf Label Flipped}  \\ \midrule
    \multicolumn{4}{c}{\bf \trainSet Set (w/o \neutral)} \\ \midrule
     0 (w/o Ino) &\bf 94.38 & 78.53 & 48.70 \\ \hdashline
     100  & 91.82 & 84.61 & 57.62 \\
     200  & 92.46 &\bf 84.92 &\bf 59.43 \\
     300  & 91.08 & 83.54 & 63.58 \\

     \midrule
      \multicolumn{4}{c}{\bf \alphaOne Set (w/o \neutral)} \\ \midrule
     0 (w/o Ino) &\bf 71.99  & 69.65  & 44.01 \\ \hdashline
     100  & 66.05 & 75.03 & 50.40 \\
     200  & 65.86 &\bf 75.03 & 50.57 \\
     300  & 65.59 & 74.23 &\bf 52.09 \\
     \bottomrule     
     \end{tabular}}
     \captionof{table}{\small Performance after inoculation fine-tuning on the  \datasetName counterfactual example sets.}
    \label{tab:counterfactual_inoculation_perturb}
 \end{minipage}

\section{Discussion and Related Work}
\label{sec:discussion}

\paragraph{What did we learn?} Firstly, through systematic probing, we have shown that despite good performance on the evaluation sets, the model for tabular NLI fails at reasoning. From the analysis of hypothesis perturbations (\S\ref{sec:probe2}), we show that the model heavily relies on correlations between a hypothesis' sentence structure and its label. Models should be systematically evaluated on adversarial sets like \alphaTwo for robustness and sensitivity. This observation is concordant with multiple studies that probe deep learning models on adversarial examples in a variety of tasks such as question answering, sentiment analysis, document classification, natural language inference, etc. \cite[e.g.][]{ribeiro2020beyond, richardson2020probing, goel2021robustness, lewis2021question, tarunesh2021trusting}.

Secondly, the model does not look at correct evidence required for reasoning, as is evident from the evidence-selection probing (\S\ref{sec:probe1}). Rather, it leverages spurious patterns and statistical correlations to make predictions. A recent study by \citet{lewis2021question} on question-answering shows that models indeed leverage spurious patterns to answer a large fraction (60-70$\%$) of questions.

Thirdly, from counterfactual probes (\S\ref{sec:probe3}), we found that the model relies on knowledge of pre-trained language models than on tabular evidence as the primary source of knowledge for making predictions. This is in addition to the spurious patterns or hypothesis artifacts leveraged by the model. Similar observations are made by~\citet{clark2016my,jia-liang-2017-adversarial, kaushik2019learning, huang2020counterfactually, gardner2020evaluating, tu2020empirical, liu2021probing, zhang2021double, wang2021logic} for unstructured text. 

Finally, from the inoculation study (\S\ref{sec:inoculation}), we found that fine-tuning on challenge sets improves model performance on challenge sets but degrades on the original \alphaOne, \alphaTwo, and \alphaThree test sets. That is, changes in the data distribution during training have a negative impact on model performance. This adds weight to the argument that the model relies excessively on data artifacts.

\paragraph{Benefits of Tabular data} Unlike unstructured data, where creating challenge datasets may be more difficult~\cite[e.g.][]{ribeiro2020beyond, goel2021robustness, mishra-etal-2021-looking}, we can analyze semi-structured  data more effectively. Although connected with the title, the rows in the table are still independent, linguistically and otherwise. Thus, controlled experiments are easier to design and study. For example, the analysis done for evidence selection via multiple table perturbation operations such as row deletion and insertion is possible mainly due to the tabular nature of the data. Such granularity and component-independence is generally absent for raw text at the token, sentence and even paragraph level. As a result, designing suitable probes with sufficient coverage can be a challenging task, and can require more manual effort. 

Additionally, probes defined on one tabular dataset (\datasetName in our case) can be easily ported to other tabular datasets such as WikiTableQA \cite{pasupat2015compositional}, TabFact \cite{chen2019tabfact}, HybridQA \cite{chen2020hybridqa, zayats2021representations, oguz2020unified}, OpenTableQA \cite{chen2020open}, ToTTo \cite{parikh2020totto}, Turing Tables \cite{yoran2021turning}, LogicTable~\cite{chen2020logical}. Moreover, such probes can be used to better understand the behavior of various tabular reasoning models~\cite[e.g.,][and others]{muller2021tapas, herzig-etal-2020-tapas,yin-etal-2020-tabert,iida-etal-2021-tabbie,pramanick2021joint,glass-etal-2021-capturing}.

\paragraph{Interpretability for NLI models} For classification tasks such as NLI, correct predictions do not always mean that the underlying model is employing correct reasoning. More work is needed to make models interpretable, either through explanations or by pointing to the evidence that is used for predictions ~\cite[e.g.][]{feng2018pathologies,serrano2019attention,jain2019attention,wiegreffe2019attention, deyoung-etal-2020-eraser, paranjape2020information, hewitt2019designing, richardson2020does, niven-kao-2019-probing,ravichander2021probing}. Many recent shared tasks on reasoning over semi-structured tabular data (such as SemEval 2021 Task 9 \cite{semeval_2021:task9} and FEVEROUS \cite{aly2021feverous}) have highlighted the importance of, and the challenges associated with, evidence extraction for claim verification.

Finally, NLI models should be tested on multiple test sets in adversarial settings \citep[e.g.,][]{ribeiro2016should, ribeiro2018anchors, ribeiro2018semantically, zhao2018generating, iyyer2018adversarial,glockner-etal-2018-breaking, naik2018stress, mccoy2019right, nie2019analyzing, liu2019inoculation} focusing on particular properties or aspects of reasoning, such as perturbed premises for evidence selection, zero-shot transfer (\alphaThree), counterfactual premises or alternate facts, and contrasting hypotheses via perturbation (\alphaTwo). Such behavioral probing by evaluating on multiple test-only benchmarks and controlled probes is essential to better understand both the abilities and the weaknesses of pre-trained language models.
\section{Conclusions}
\label{sec:conclusion}
This paper presented a targeted probing study to highlight the limitations of tabular inference models using a case study on a tabular NLI task on \datasetName. Our findings show that despite good performance on standard splits, a RoBERTa-based tabular NLI model, fine-tuned on the existing pre-trained language model, fails to select the correct evidence, makes incorrect predictions on adversarial hypotheses, and is not grounded in provided evidence--counterfactual or otherwise. We expect that insights from the study can help in designing rationale selection techniques based on structural constraints for tabular inference and other tasks. While inoculation experiments showed partial success, diverse data augmentation may help mitigate challenges. However, annotation of such data can be expensive. It may also be possible to train models to satisfy domain-based constraints~\cite[e.g.,][]{li2020structured} to improve model robustness. Finally, probing techniques described here may be adapted to other NLP tasks involving tables such as tabular question answering and tabular text generation.
\section*{Acknowledgements}
\label{sec:aknowledgement}
We thank the reviewing team for their valuable feedback. This work is partially supported by NSF grants \#1801446 and \#1822877.

\bibliography{custom,anthology}
\bibliographystyle{acl_natbib}

\appendix

\section*{Appendix}
\label{app:appendix-automatic-probing}
In Section \ref{sec:probe1automatic}, we defined four types of row-agnostic table modifications:(a) row deletion, (b) row insertion, (c) row-value update,  and (d) row permutation and presented the first one there. We present details of the rest here along with the respective impact on the \alphaOne, \alphaTwo and \alphaThree test sets.

\paragraph{Row Insertion.}
When we insert new information that does not contradict an existing table,\footnote{To ensure that the information added is not contradictory to existing rows, we only add rows with new keys instead of changing values for the existing keys.} original predictions should be retained in almost all cases. Very rarely, \neutral labels may change to \entail or \contradict. For example, adding the \textit{Singles} row below to our running example table doesn't change labels for any hypothesis except the H4 label (see Figure \ref{fig:example}) changing to \contradict with the additional information.

\begin{minipage}{.9\columnwidth}
\bigskip  
\small
    \begin{tabular}{>{\raggedright}p{0.2\linewidth}p{0.7\linewidth}}
    \toprule
     \bf Singles & The Logical Song; Breakfast in America; Goodbye Stranger; Take the Long Way Home \\
     \bottomrule
    \end{tabular}
\bigskip
\end{minipage}

Figure \ref{fig:probe1insertion} shows the possible label changes after new row insertion as a directed labeled graph, and the results are summarized in Table \ref{tab:summary_rowinsertion}. Note that all transitions from \neutral are valid upon row insertion, although not all may be accurate.

\begin{minipage}{0.93\columnwidth}
\bigskip
    \begin{tikzpicture}[thick, scale=0.6, {every node/.style}={scale=0.6}]
\begin{pgfonlayer}{nodelayer}
		\node [style=new style 0] (0) at (0, 8) {C};
		\node [style=new style 0] (1) at (-7, 0) {E};
		\node [style=new style 0] (2) at (7, 0) {N};
		\node [style=none] (3) at (-6.5, 1.5) {};
		\node [style=none] (4) at (-6, 1) {};
		\node [style=none] (5) at (-5.75, -0.5) {};
		\node [style=none] (6) at (-5.75, 0.25) {};
		\node [style=none] (7) at (-1.25, 7.25) {};
		\node [style=none] (8) at (-0.75, 6.75) {};
		\node [style=none] (9) at (0.75, 6.75) {};
		\node [style=none] (10) at (1.25, 7.25) {};
		\node [style=none] (11) at (6, 1) {};
		\node [style=none] (12) at (6.5, 1.5) {};
		\node [style=none] (13) at (5.5, -0.5) {};
		\node [style=none] (14) at (5.5, 0.25) {};
		\node [style=new style 1] (20) at (0, 11.75) {93.23, 93.46, 93.65};
		\node [style=new style 1] (21) at (4.25, 7.75) {};
		\node [style=new style 1] (22) at (7.5, -3.5) {94.39, 95.16, 92.66};
		\node [style=new style 1] (23) at (0, 1) {3.05, 2.79, 3.87};
		\node [style=new style 1] (24) at (-7.25, -3.75) {97.19, 95.01, 97.48};
		\node [style=new style 2] (29) at (-7, 4) {1.58, 3.26, 1.28};
		\node [style=new style 2] (30) at (-5, 8) {4.32, 4.32, 4.16};
		\node [style=new style 2] (33) at (0, -1.25) {1.23, 1.73, 1.23};
		\node [style=none] (34) at (-5, 7.5) {};
		\node [style=none] (35) at (-2.5, 4.75) {};
		\node [style=none] (36) at (3, 4.25) {};
		\node [style=none] (37) at (4.25, 7.25) {};
		\node [style=none] (39) at (-7.75, -1.5) {*};
		\node [style=new style 1] (40) at (6.5, 4) {};
		\node [style=new style 1] (41) at (7, 4) {2.56, 2.05, 3.47};
		\node [style=new style 2] (42) at (4.25, 7.75) {};
		\node [style=new style 2] (43) at (4.25, 7.75) {2.45, 2.22, 2.19};
		\node [style=none] (44) at (-0.5, 9) {*};
	\end{pgfonlayer}
	\begin{pgfonlayer}{edgelayer}
		\draw [style=new edge style 0] (3.center) to (7.center);
		\draw [style=new edge style 0] (8.center) to (4.center);
		\draw [style=new edge style 5, bend left=45, looseness=1.75] (35.center) to (34.center);
		\draw [style=new edge style 5, bend right=45, looseness=1.25] (36.center) to (37.center);
		\draw [style=new edge style 0] (5.center) to (13.center);
		\draw [style=new edge style 4] (14.center) to (6.center);
		\draw [style=new edge style 0] (9.center) to (11.center);
		\draw [style=new edge style 1] (10.center) to (12.center);
		\draw [style=new edge style 3, in=-135, out=-45, loop] (1) to ();
		\draw [style=new edge style 3, in=135, out=45, loop] (0) to ();
		\draw [style=new edge style 5, in=-135, out=-45, loop] (2) to ();
	\end{pgfonlayer}
\end{tikzpicture}
    \captionof{figure}{\small Changes in model predictions after new row insertion. (Notation similar to Figure \ref{fig:probe1deletion})}
    \label{fig:probe1insertion}
\end{minipage}

\begin{minipage}{0.93\columnwidth}
\bigskip
\small
    \begin{tabular}{l|c|c|c|c}
    \toprule
    Dataset & \alphaOne & \alphaTwo & \alphaThree & Average \\ \midrule
        \entail & 2.81 & 4.99  & 2.51 & 3.44\\
        \neutral & 0 & 0 & 0 & 0\\
        \contradict & 6.77 & 6.54 & 6.35 & 6.55\\ \midrule
        Average & 3.19 & 3.84 & 2.95 & - \\
        \bottomrule
    \end{tabular}
    \captionof{table}{\small Percentage of invalid transitions after new row insertion. For an ideal model, all these numbers should be zero.}
    \label{tab:summary_rowinsertion}
\smallskip
\end{minipage}

\paragraph{Row Update.}
In case of row update, we only change a portion of a row value. Whole row value substitutions are examined separately as composite operations of deletion followed by insertion. Unlike a whole row update, changing only a portion of a row is non-trivial. We must ensure that the updated value is appropriate for the key in question and also avoid self-contradictions. To satisfy these constraints, we update a row with a value from a random table with the same key and only update values in multi-valued rows. A row update operation may have an effect on all labels. Though feasible, we consider the transitions from \contradict to \entail to be prohibited. Unlike \entail to \contradict transitions, these transitions would be extremely rare as values are updated randomly, regardless of their semantics. For example, if we substitute \textit{pop} in the multi-valued key \rowKey{Genre} in our running example with another genre, the hypothesis H1 is likely to change to \contradict. 

\begin{minipage}{0.93\columnwidth}
\bigskip
    \begin{tikzpicture}[thick, scale=0.6, {every node/.style}={scale=0.6}]
	\begin{pgfonlayer}{nodelayer}
		\node [style=new style 0] (0) at (0, 8) {C};
		\node [style=new style 0] (1) at (-7, 0) {E};
		\node [style=new style 0] (2) at (7, 0) {N};
		\node [style=none] (3) at (-6.5, 1.5) {};
		\node [style=none] (4) at (-6, 1) {};
		\node [style=none] (5) at (-5.75, -0.5) {};
		\node [style=none] (6) at (-5.75, 0.25) {};
		\node [style=none] (7) at (-1.25, 7.25) {};
		\node [style=none] (8) at (-0.75, 6.75) {};
		\node [style=none] (9) at (0.75, 6.75) {};
		\node [style=none] (10) at (1.25, 7.25) {};
		\node [style=none] (11) at (6, 1) {};
		\node [style=none] (12) at (6.5, 1.5) {};
		\node [style=none] (13) at (5.5, -0.5) {};
		\node [style=none] (14) at (5.5, 0.25) {};
		\node [style=new style 1] (20) at (0, 11.75) {99.51, 99.69, 99.81};
		\node [style=new style 1] (22) at (7.5, -3.75) {99.63, 99.69, 99.9};
		\node [style=new style 1] (23) at (0, 1) {};
		\node [style=new style 1] (24) at (-7.25, -3.75) {99.7, 99.24, 99.63};
		\node [style=new style 1] (29) at (-8, 4) {0.23, 0.53, 0.25};
		\node [style=new style 2] (30) at (-5, 8) {0.3, 0.21, 0.09};
		\node [style=new style 2] (33) at (0, -1.25)  {0.08, 0.22, 0.12} ;
		\node [style=none] (34) at (-5, 7.5) {};
		\node [style=none] (35) at (-2.5, 4.75) {};
		\node [style=none] (36) at (3, 4.25) {};
		\node [style=none] (37) at (5.25, 7.25) {};
		\node [style=new style 2] (43) at (5.5, 7.75) {0.19, 0.09, 0.10};
		\node [style=none] (45) at (8, -1.5) {*};
		\node [style=new style 1] (47) at (8, 4) {0.26, 0.20, 0.02};
		\node [style=new style 2] (48) at (0, 1) {0.12, 0.11, 0.07} ;
	\end{pgfonlayer}
	\begin{pgfonlayer}{edgelayer}
		\draw [style=new edge style 0] (8.center) to (4.center);
		\draw [style=new edge style 5, bend left=45, looseness=1.75] (35.center) to (34.center);
		\draw [style=new edge style 5, bend right=45, looseness=1.25] (36.center) to (37.center);
		\draw [style=new edge style 0] (5.center) to (13.center);
		\draw [style=new edge style 0] (9.center) to (11.center);
		\draw [style=new edge style 0] (14.center) to (6.center);
		\draw [style=new edge style 3, in=-135, out=-45, loop] (2) to ();
		\draw [style=new edge style 1] (7.center) to (3.center);
		\draw [style=new edge style 1] (10.center) to (12.center);
		\draw [style=new edge style 2, in=135, out=45, loop] (0) to ();
		\draw [style=new edge style 2, in=-135, out=-45, loop] (1) to ();
	\end{pgfonlayer}
\end{tikzpicture}
    \captionof{figure}{\small Changes in model predictions after row value update. (Notation similar to Figure \ref{fig:probe1deletion})}
    \label{fig:probe1update}
    \bigskip
\end{minipage}

Since we are updating a single value from a multi-valued key, the changes to the table are minimal and may not be perceived by the model. As a result, we should expect row updates to have lower impact on model predictions. This appears to be the case, as evidenced by the results in Figure \ref{fig:probe1update}, which show that the labels do not change drastically after update. The results in Figure \ref{fig:probe1update} are summarized in Table \ref{tab:summary_rowupdate}.

\begin{minipage}{0.93\columnwidth}
\bigskip
\small
\begin{tabular}{p{2cm}|p{0.64cm}|p{.64cm}|p{0.64cm}|p{0.9cm}}
\toprule
    Dataset & \alphaOne & \alphaTwo & \alphaThree & Average \\ \midrule
        \entail & 0.08 & 0.22  & 0.12 & 0.14\\
        \neutral & 0.12 & 0.11  & 0.09 & 0.11\\
        \contradict & 0.49 & 0.30 & 0.19 & 0.33\\ \midrule
        Average & 0.23 & 0.21 & 0.13 &  -\\
        \bottomrule
    \end{tabular}
    \captionof{table}{\small Percentage of invalid transitions after row value update. For an ideal model, all these numbers should be zero.}
    \label{tab:summary_rowupdate}
\smallskip
\end{minipage}

\paragraph{Row Permutation.} By design of the premises, the order of their rows should have no effect on hypotheses labels. In other words, the labels should be invariant to row permutation. However, from Figure \ref{fig:probe1perturb}, it is evident that even a simple shuffling of rows, where no information has been tampered with, can have a notable effect on performance. This shows that the model is relying on row positions incorrectly, while the semantics of a table is order invariant. We summarize the combined invalid transitions from Figure \ref{fig:probe1perturb} in Table \ref{tab:summary_rowpermutation}.

\begin{minipage}{0.93\columnwidth}
\bigskip
    \begin{tikzpicture}[thick, scale=0.6, {every node/.style}={scale=0.6}]
	\begin{pgfonlayer}{nodelayer}
		\node [style=new style 0] (0) at (0, 8) {C};
		\node [style=new style 0] (1) at (-7, 0) {E};
		\node [style=new style 0] (2) at (7, 0) {N};
		\node [style=none] (3) at (-6.5, 1.5) {};
		\node [style=none] (4) at (-6, 1) {};
		\node [style=none] (5) at (-5.75, -0.5) {};
		\node [style=none] (6) at (-5.75, 0.25) {};
		\node [style=none] (7) at (-1.25, 7.25) {};
		\node [style=none] (8) at (-0.75, 6.75) {};
		\node [style=none] (9) at (0.75, 6.75) {};
		\node [style=none] (10) at (1.25, 7.25) {};
		\node [style=none] (11) at (6, 1) {};
		\node [style=none] (12) at (6.5, 1.5) {};
		\node [style=none] (13) at (5.5, -0.5) {};
		\node [style=none] (14) at (5.5, 0.25) {};
		\node [style=new style 1] (20) at (0, 11.75) {88.37, 91.25, 86.30};
		\node [style=new style 1] (22) at (7.5, -3.75) {92.94, 93.10, 87.55};
		\node [style=new style 1] (23) at (0, 1) {};
		\node [style=new style 1] (24) at (-7.25, -3.75) {90.75, 87.74, 85.41};
		\node [style=new style 2] (30) at (-5, 8) {6.95, 5.76, 6.91};
		\node [style=new style 2] (33) at (0, -1.25) {3.47, 4.82, 6.82};
		\node [style=none] (34) at (-5, 7.5) {};
		\node [style=none] (35) at (-2.5, 4.75) {};
		\node [style=none] (36) at (3, 4.25) {};
		\node [style=none] (37) at (5.25, 7.25) {};
		\node [style=new style 2] (43) at (5.5, 7.75) {4.65, 2.96, 6.85};
		\node [style=none] (45) at (8, -1.5) {*};
		\node [style=new style 2] (47) at (8, 4) {};
		\node [style=new style 2] (48) at (0, 1) {3.24, 3.78, 5.84};
		\node [style=new style 1] (50) at (8, 4) {};
		\node [style=none] (51) at (-6.75, -1.5) {*};
		\node [style=none] (52) at (0.5, 9) {*};
		\node [style=new style 2] (53) at (8, 4) {};
		\node [style=new style 2] (54) at (8, 4) {3.82, 3.12, 6.61};
		\node [style=new style 2] (56) at (-6.75, 4) {5.78, 7.44, 7.77};
	\end{pgfonlayer}
	\begin{pgfonlayer}{edgelayer}
		\draw [style=new edge style 0] (8.center) to (4.center);
		\draw [style=new edge style 5, bend left=45, looseness=1.75] (35.center) to (34.center);
		\draw [style=new edge style 5, bend right=45, looseness=1.25] (36.center) to (37.center);
		\draw [style=new edge style 0] (5.center) to (13.center);
		\draw [style=new edge style 0] (9.center) to (11.center);
		\draw [style=new edge style 0] (14.center) to (6.center);
		\draw [style=new edge style 3, in=-135, out=-45, loop] (2) to ();
		\draw [style=new edge style 2, in=135, out=45, loop] (0) to ();
		\draw [style=new edge style 0] (3.center) to (7.center);
		\draw [style=new edge style 0] (12.center) to (10.center);
		\draw [style=new edge style 1, in=135, out=45, loop] (0) to ();
		\draw [style=new edge style 1, in=-135, out=-45, loop] (1) to ();
	\end{pgfonlayer}
\end{tikzpicture}
    \captionof{figure}{\small Changes in model predictions after shuffling of table rows. (Notation similar to Figure \ref{fig:probe1deletion}.)}
    \label{fig:probe1perturb}
\end{minipage}

\begin{minipage}{0.93\columnwidth}
\bigskip
\small
  \captionsetup{justification=raggedright}
    \begin{tabular}{p{2cm}|p{0.64cm}|p{.64cm}|p{0.64cm}|p{0.9cm}}
    \toprule
        Dataset & \alphaOne & \alphaTwo & \alphaThree & Average\\ \midrule
        \entail & 9.25 & 12.2 & 14.6 & 12.02\\
        \neutral & 7.1 & 6.8  & 12.5 & 8.79\\
        \contradict & 11.6 & 8.76  & 13.7 & 11.36\\ \midrule
        Average & 9.34 & 9.26 & 13.6 & -\\
        \bottomrule
    \end{tabular}
    \captionof{table}{\small Percentage of invalid transitions after row permutations. For an ideal model, all these numbers should be zero.}
    \label{tab:summary_rowpermutation}
\end{minipage}

\paragraph{Irrelevant Row Deletion.} Ideally, deletion of an irrelevant row should have no effect on a hypothesis label. The results in Figure \ref{fig:probe1deletionirrelevant} and in Table \ref{tab:summary_rowirrrelevant_deletion} show that even irrelevant rows have an effect on model predictions. This further illustrates that the seemingly accurate model predictions are not appropriately grounded on evidence.

\begin{minipage}{\columnwidth}
\bigskip
    \begin{tikzpicture}[thick, scale=0.6, {every node/.style}={scale=0.6}]
	\begin{pgfonlayer}{nodelayer}
		\node [style=new style 0] (0) at (0, 8) {C};
		\node [style=new style 0] (1) at (-7, 0) {E};
		\node [style=new style 0] (2) at (7, 0) {N};
		\node [style=none] (3) at (-6.5, 1.5) {};
		\node [style=none] (4) at (-6, 1) {};
		\node [style=none] (5) at (-5.75, -0.5) {};
		\node [style=none] (6) at (-5.75, 0.25) {};
		\node [style=none] (7) at (-1.25, 7.25) {};
		\node [style=none] (8) at (-0.75, 6.75) {};
		\node [style=none] (9) at (0.75, 6.75) {};
		\node [style=none] (10) at (1.25, 7.25) {};
		\node [style=none] (11) at (6, 1) {};
		\node [style=none] (12) at (6.5, 1.5) {};
		\node [style=none] (13) at (5.5, -0.5) {};
		\node [style=none] (14) at (5.5, 0.25) {};
		\node [style=new style 1] (20) at (0, 11.75) {};
		\node [style=new style 1] (22) at (7.5, -3.75) {96.12, 96.46, 94.99};
		\node [style=new style 1] (23) at (0, 1) {};
		\node [style=new style 1] (24) at (-7.25, -3.75) {};
		\node [style=new style 2, rotate=47] (29) at (-4.75, 4.5) {2.76, 4.05, 3.46};
		\node [style=new style 2, rotate=47] (30) at (-3, 3.25) {2.93, 2.94, 2.94};
		\node [style=new style 2] (33) at (0, -1.25) {};
		\node [style=new style 2] (48) at (0, 1) {1.85, 1.62, 2.32};
		\node [style=new style 2] (49) at (0, -1.25) {};
		\node [style=new style 2] (50) at (0, -1.25) {2.38, 2.92, 2.63};
		\node [style=new style 1] (51) at (0, 11.75) {};
		\node [style=new style 1] (52) at (0, 11.75) {94.06, 94.92 , 93.09};
		\node [style=new style 1] (53) at (-7.25, -3.75) {94.86, 93.03, 93.91};
		\node [style=new style 2, rotate=-47] (56) at (4.25, 5) {2.05, 1.92, 2.69};
		\node [style=new style 2, rotate=-47] (57) at (2.75, 3.5) {3.01, 2.15, 3.96};
	\end{pgfonlayer}
	\begin{pgfonlayer}{edgelayer}
		\draw [style=new edge style 0] (8.center) to (4.center);
		\draw [style=new edge style 0] (5.center) to (13.center);
		\draw [style=new edge style 0] (9.center) to (11.center);
		\draw [style=new edge style 0] (14.center) to (6.center);
		\draw [style=new edge style 3, in=-135, out=-45, loop] (2) to ();
		\draw [style=new edge style 2, in=135, out=45, loop] (0) to ();
		\draw [style=new edge style 0] (3.center) to (7.center);
		\draw [style=new edge style 0] (12.center) to (10.center);
		\draw [style=new edge style 1, in=135, out=45, loop] (0) to ();
		\draw [style=new edge style 1, in=-135, out=-45, loop] (1) to ();
	\end{pgfonlayer}
\end{tikzpicture}
    \captionof{figure}{\small Change in model predictions after deletion of an \textbf{irrelevant row}. (Notation similar to Figure \ref{fig:probe1deletion}.)}
    \label{fig:probe1deletionirrelevant}
\end{minipage}

\begin{minipage}{0.93\columnwidth}
\bigskip
    \small
    \centering
    \begin{tabular}{l|c|c|c|c}
    \toprule
        Datasets & \alphaOne & \alphaTwo & \alphaThree & Average \\ \midrule
        \entail & 5.14 & 6.97 &	6.09 & 6.07	\\
        \neutral & 3.9	& 3.54 & 5.01 & 4.15 \\
        \contradict & 5.94 & 5.09 & 6.91 & 5.98 \\
        \midrule
        Average & 4.99 & 5.2 & 6.01 & -\\
        \bottomrule
    \end{tabular}
    \captionof{table}{\small Percentage of invalid transitions after deletion of irrelevant rows. For an ideal model, all these numbers should be zero.}
    \label{tab:summary_rowirrrelevant_deletion}
\end{minipage}

\paragraph{Composition of Perturbation Operations}
\label{sec:appendix_compositionality}
In addition to probing individual operations, we can also study their compositions. For example, we could delete a row, and insert a different row, and so on. The composition of these operations have interesting properties with respect to the allowed transitions. For example, when an operation is composed with itself (e.g. two deletions), the set of valid label changes is the same as for the operation. A particularly interesting composition is deletion followed by an insertion, since this can be viewed as a row update. In Figure \ref{fig:probecompositiondeleteinsert}, we show the transition graph for the composition operation of row deletion followed by insertion and the summary of the possible transitions is presented in Table \ref{tab:summary_compositiondeleteinsert}.

\begin{minipage}{0.93\columnwidth}
\bigskip
\centering
    \begin{tikzpicture}[thick, scale=0.6, {every node/.style}={scale=0.6}]
	\begin{pgfonlayer}{nodelayer}
		\node [style=new style 0] (0) at (0, 8) {C};
		\node [style=new style 0] (1) at (-7, 0) {E};
		\node [style=new style 0] (2) at (7, 0) {N};
		\node [style=none] (3) at (-6.5, 1.5) {};
		\node [style=none] (4) at (-6, 1) {};
		\node [style=none] (5) at (-5.75, -0.5) {};
		\node [style=none] (6) at (-5.75, 0.25) {};
		\node [style=none] (7) at (-1.25, 7.25) {};
		\node [style=none] (8) at (-0.75, 6.75) {};
		\node [style=none] (9) at (0.75, 6.75) {};
		\node [style=none] (10) at (1.25, 7.25) {};
		\node [style=none] (11) at (6, 1) {};
		\node [style=none] (12) at (6.5, 1.5) {};
		\node [style=none] (13) at (5.5, -0.5) {};
		\node [style=none] (14) at (5.5, 0.25) {};
		\node [style=new style 1] (20) at (0, 11.75) {};
		\node [style=new style 1] (22) at (7.5, -3.75) {86.67, 87.46, 86.35};
		\node [style=new style 1] (23) at (0, 1) {};
		\node [style=new style 2] (24) at (-7.25, -3.75) {};
		\node [style=new style 2, rotate=47] (29) at (-4.75, 4.5) {3.02, 6.53, 4.16};
		\node [style=new style 2, rotate=47] (30) at (-3, 3.25) {9.81, 7.88, 6.72};
		\node [style=new style 2] (33) at (0, -1.25) {};
		\node [style=new style 1] (48) at (0, 1) {7.47, 7.15, 6.11};
		\node [style=new style 2] (49) at (0, -1.25) {};
		\node [style=new style 1] (50) at (0, -1.25) {1.76, 2.07, 3.28};
		\node [style=new style 1] (51) at (0, 11.75) {};
		\node [style=new style 1] (52) at (0, 11.75) {86.03, 89.19, 89.13};
		\node [style=new style 1] (53) at (-7.25, -3.75) {95.42, 91.40, 92.56};
		\node [style=new style 1, rotate=-47] (56) at (4.25, 5) {5.66, 5.39. 7.53};
		\node [style=new style 1, rotate=-47] (57) at (2.75, 3.5) {4.16, 2.93, 4.15};
	\end{pgfonlayer}
	\begin{pgfonlayer}{edgelayer}
		\draw [style=new edge style 0] (8.center) to (4.center);
		\draw [style=new edge style 1] (6.center) to (14.center);
		\draw [style=new edge style 1, in=-135, out=-45, loop] (2) to ();
		\draw [style=new edge style 1] (13.center) to (5.center);
		\draw [style=new edge style 0] (3.center) to (7.center);
		\draw [style=new edge style 1] (11.center) to (9.center);
		\draw [style=new edge style 1] (10.center) to (12.center);
		\draw [style=new edge style 1, in=-135, out=-45, loop] (1) to ();
		\draw [style=new edge style 1, in=135, out=45, loop] (0) to ();
	\end{pgfonlayer}
\end{tikzpicture}
    \captionof{figure}{\small Changes in model predictions after deletion followed by an insert operation. (Notation similar to Figure \ref{fig:probe1deletion}.)}
    \label{fig:probecompositiondeleteinsert}
    \bigskip
\end{minipage}

\begin{minipage}{0.93\columnwidth}
\bigskip
    \small
    \centering
    \begin{tabular}{l|c|c|c|c}
    \toprule
        Datasets & \alphaOne & \alphaTwo & \alphaThree & Average \\ \midrule
        \entail & 3.02 & 6.53 & 4.16 & 4.57	\\
        \neutral & 0.00	& 0.00 & 0.00 & 0.00 \\
        \contradict & 9.81 & 7.88 & 6.71 & 8.13 \\
        \midrule
        Average & 4.28 & 4.80 & 3.63 & -\\
        \bottomrule
    \end{tabular}
    \captionof{table}{\small Percentage of invalid transitions after deletion followed by an insertion operation. For an ideal model, all these numbers should be zero.}
    \label{tab:summary_compositiondeleteinsert}
\end{minipage}

\end{document}